\documentclass[journal]{IEEEtran}
\usepackage{amsmath,amsfonts}
\usepackage{algorithmic}
\usepackage{algorithm}
\usepackage{array}
\usepackage[caption=false,font=normalsize,labelfont=sf,textfont=sf]{subfig}
\usepackage{textcomp}
\usepackage{stfloats}
\usepackage{url}
\usepackage{verbatim}
\usepackage{graphicx}
\usepackage{booktabs} 
\usepackage{cite}
\usepackage[dvipsnames]{xcolor}
\usepackage{lipsum}
\usepackage{placeins}

\makeatletter
\let\NAT@parse\undefined
\makeatother
\usepackage{hyperref}

\hypersetup{
    colorlinks,
    linkcolor={red!50!black},
    citecolor={red!50!black},
    urlcolor={blue!80!black}
}

\IEEEoverridecommandlockouts    

\title{\LARGE \bf{Evaluating Text-to-Image Diffusion Models \\ for Texturing Synthetic Data}}

\author{Thomas Lips$^{1}$ and  Francis wyffels$^{1}$
\thanks{$^{1}$AI and Robotics Lab, Ghent University - imec \newline Technologiepark 126, 9052 Zwijnaarde, Belgium}
\thanks{corresponding author: Thomas.Lips@UGent.be}
}

\begin{document}

\maketitle
\thispagestyle{empty} 
\pagestyle{empty}

\begin{abstract}
Building generic robotic manipulation systems often requires large amounts of real-world data, which can be dificult to collect. Synthetic data generation offers a promising alternative, but limiting the sim-to-real gap requires significant engineering efforts. To reduce this engineering effort, we investigate the use of pretrained text-to-image diffusion models for texturing synthetic images and compare this approach with using random textures, a common domain randomization technique in synthetic data generation. We focus on generating object-centric representations, such as keypoints and segmentation masks, which are important for robotic manipulation and require precise annotations. We evaluate the efficacy of the texturing methods by training models on the synthetic data and measuring their performance on real-world datasets for three object categories: shoes, T-shirts, and mugs. Surprisingly, we find that texturing using a diffusion model performs on par with random textures, despite generating seemingly more realistic images. Our results suggest that, for now, using diffusion models for texturing does not benefit synthetic data generation for robotics. 

The code, data and trained models are available at \url{https://github.com/tlpss/diffusing-synthetic-data.git}.


\end{abstract}

\section{Introduction}
Building generic robotic manipulation systems often requires learning-based methods to deal with the diversity in environments and objects. The performance of these learned models largely depends on the amount of data available to train them. Real-world (robot) data collection is time-consuming~\cite{manuelli2019kpam,levine2018learning,lynch2023languagetable}. Therefore, the amount of data is often a bottleneck to the creation of generic robotic manipulation systems. Pretrained foundation models can reduce this need for task-specific data, but they cannot yet cover all use cases~\cite{foundation_models_review}.

A parallel approach to overcome this data bottleneck is to train the robot system on synthetically generated data instead of real-world data. 
The main difficulty with synthetic data is to make sure that models trained on this synthetic data transfer well to the real-world, i.e., to limit the sim-to-real performance gap~\cite{syntheticreview}. In practice, this often requires  significant amounts of manual engineering for  3D asset generation, scene composition and texturing.\cite{hodavn2019photorealistic,wood2021fakeit,plum2023replicant,andrychowicz2020openaiRubiks, lips2024cloth-keypoints}.
 In this work we focus on texturing, which can be summarized as creating the appearance of 3D objects and scenes by specifying the optical properties (color being the most important one) of each part of an object. Recently, researchers have sought to (partially) outsource this work to neural networks, for example, by generating synthetic images using text-to-image diffusion models~\cite{yu2023rosie,chen2023genaug,ma2023diffusionseg}. 

\begin{figure*}
    \centering
    \includegraphics[width=1.0\linewidth]{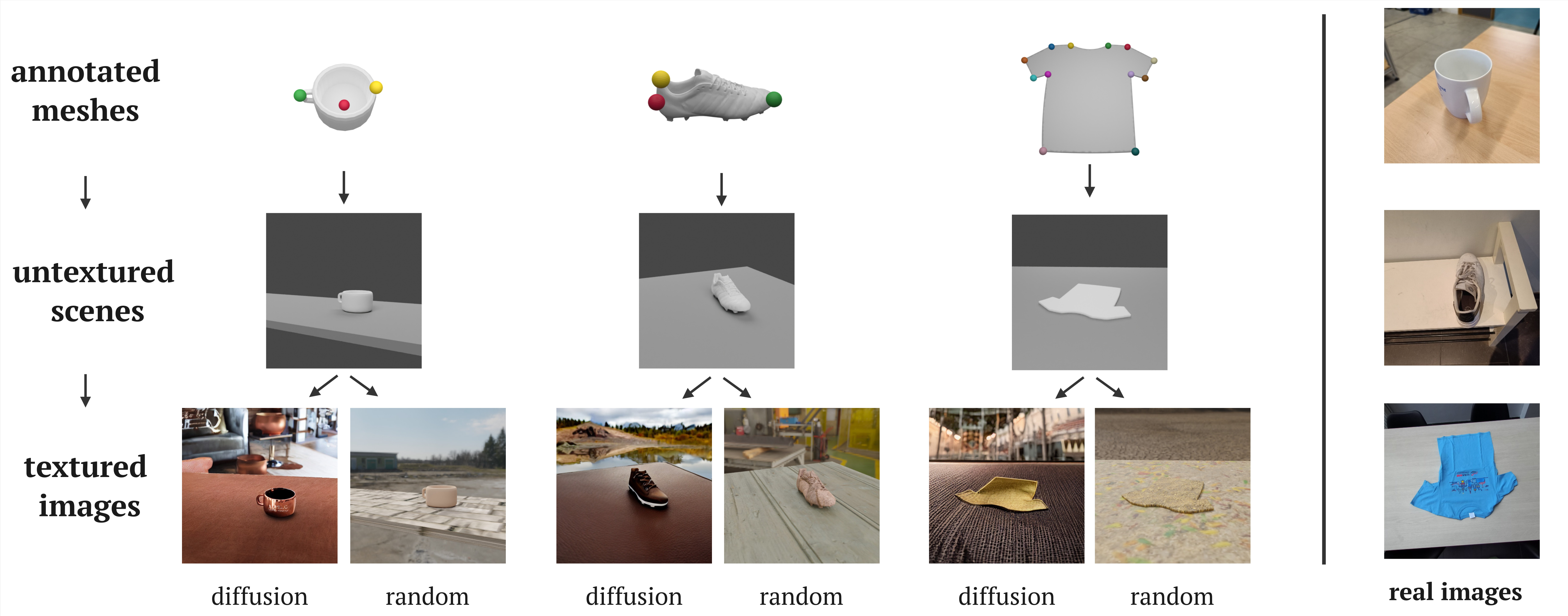}
    \caption{Left: In this work, we compare text-to-image diffusion models against random textures for texturing 3D scenes in a synthetic data generation pipeline. Right: We  evaluate the efficacy of the synthetic data on real-world data for both keypoint detection and segmentation.}
    \label{fig:overview}
\end{figure*}

In this work, we further investigate the use of text-to-image diffusion models to texture RGB images of a 3D scene and compare this method against using random textures. We also explore various design choices and  pipelines for diffusion-based texturing. To generate the synthetic images, we first create a 3D scene and obtain the annotations for that scene. We then texture the scene by either adding random textures to all elements or by using a diffusion model to generate  the textures. The process is illustrated in Figure~\ref{fig:overview}.

We focus on pixel-level representations that are often used in robotic manipulation: keypoints and segmentation masks. These representations require precise annotations and therefore we first create an explicit 3D scene instead of directly generating images from text prompts using a text-to-image diffusion model: From the 3D scene, we can extract pixel-perfect annotations. In addition, we need to ensure that the diffusion model does not alter the semantics of the scene during texturing as this would invalidate the annotations. For example, the diffusion model cannot alter the shape of the object or change its pose in the image. To accomplish this, we use a Controlnet~\cite{zhang2023controlnet} to condition on  both a depth image of the scene and a prompt, as in~\cite{chen2023genaug, ma2024generatingimages3dannotations}.

We evaluated the efficacy of the data generation methods by measuring the downstream performance of models trained on the data. We generated data for static scenes of 3 different object categories lying on a table: shoes, T-shirts and mugs. The models were evaluated on real-world test datasets using common metrics for each representation: mean average precision (mAP) for segmentation and average keypoint distance (AKD) for keypoint detection.

Surprisingly, we found that texturing using a diffusion model performs similarly to using random textures. In a series of additional experiments, we observed that both methods exhibit limited scaling behavior and that using LLM-generated prompts resulted in the best performance for diffusion-based texturing.

To summarize, our contributions are as follows:

\begin{itemize}
    \item We are the first to use text-to-image models to generate synthetic data for keypoint detection, an important task for robotic manipulation systems which requires fine-grained annotations.
    \item We extensively compare diffusion-based synthetic data texturing against using random textures and find they perform similarly on both keypoint detection and semantic segmentation tasks.
    \item We provide insight into the use of diffusion models for synthetic data generation by analyzing the scaling behavior of both methods and evaluating a number of design choices.
\end{itemize}

\section{Related Work}
\subsection{Synthetic Data Generation}

Synthetic data generation offers a compelling alternative to manual data collection for supervised machine learning: It provides arbitrary amounts of perfectly labeled data enabling the desired generalizations. The main challenge lies in overcoming the sim-to-real gap to ensure models trained on synthetic data generalize to real-world scenarios~\cite{syntheticreview}.  Common strategies to overcome this gap include domain randomization~\cite{tremblay2018training}, in which the appearance and shape of objects, or the composition of the scene is varied beyond what is considered realistic, and domain adaptation~\cite{ho2021retinagan}, in which the differences between synthetic and real data are explicitly learned. Despite these recipes, achieving strong sim-to-real performance often requires significant human effort to improve the diversity and quality of assets (object shapes, materials) and scene compositions used for data generation ~\cite{wood2021fakeit,plum2023replicant,hodavn2019photorealistic}.

Researchers have  tried to reduce human effort using generative models. For instance, \cite{besnier2020dataset} uses a class-conditioned GAN~\cite{goodfellow2020generative} to train classifiers on images generated by the GAN. \cite{zhang2021datasetgan} goes beyond image-level semantics and trains a decoder on the latent codes of a GAN to obtain segmentation masks and keypoints for generated images automatically. However, the quality of the images generated by such GANs is limited. Furthermore, these GANs need to be explicitly trained on each category. Recently, large, pretrained text-to-image diffusion models~\cite{ddpm} have been explored to  overcome these limitations.

\subsection{Text-to-image diffusion models for synthetic data}
Text-to-image diffusion models~\cite{rombach2022stablediffusion} have been used to generate synthetic data for image classification~\cite{trabucco2023DA-Fusion,fan2023synthetic-diffusion-scaling-laws, geng2024unmetpromisesynthetictraining, ma2024generatingimages3dannotations}, semantic segmentation~\cite{nguyen2024datasetdiffusion,ma2023diffusionseg,wu2023diffumask}, 3D pose estimation~\cite{ma2024generatingimages3dannotations} and augmentation of robot trajectories~\cite{yu2023rosie,chen2023genaug}. 

For image classification, \cite{fan2023synthetic-diffusion-scaling-laws} show that diffusion-generated synthetic data does not scale as well as real data. \cite{geng2024unmetpromisesynthetictraining}~showed that directly training on the underlying dataset of the generative model can outperform training on synthetic images generated by the diffusion model. \cite{trabucco2023DA-Fusion,de2023medical}~have focused on using textual inversion~\cite{gal2022textual-inversion} to generate data of less-familiar categories, which is an important limitation of using text-to-image models for synthetic data generation.

For segmentation, a pixel-perfect object mask is required in addition to controlling the object category with a textual description. \cite{wu2023diffumask,ma2023diffusionseg,nguyen2024datasetdiffusion} use the cross-attention between text and images in  Stable Diffusion~\cite{rombach2022stable-diffusion} to generate these masks automatically. \cite{nguyen2024datasetdiffusion}~uses self-attention to further improve the generated masks and generates multi-class annotations. 

For pose estimation, \cite{ma2024generatingimages3dannotations} uses 3D meshes to generate edge maps and then renders images for these edge maps using a Controlnet. They also report results for segmentation and classification, improving on previous methods that do not use explicit 3D control. 

For data augmentation, \cite{yu2023rosie}  augments robot trajectories by inpainting parts of the image. \cite{chen2023genaug} first renders depth images of an object and then  uses a Controlnet~\cite{zhang2023controlnet} to texture them, after which they are used to augment the robot trajectories.

In ~\cite{chen2023genaug,ma2024generatingimages3dannotations}, Controlnet~\cite{zhang2023controlnet} is used to  condition on both text prompts and renders of 3D objects to increase control over the semantics of the generated images. 

Various prompt strategies have been explored for the diffusion models, including fixed templates~\cite{fan2023synthetic-diffusion-scaling-laws}, generated image captions~\cite{fan2023synthetic-diffusion-scaling-laws,wu2023diffumask} and prompts generated by LLMs~\cite{ma2024generatingimages3dannotations,ma2023diffusionseg}.


In this work, we require precise, pixel-level annotations. We, therefore, follow~\cite{chen2023genaug,ma2024generatingimages3dannotations} and condition on both text prompts and renders of 3D scenes using a Controlnet with Stable Diffusion. Our work is most related to~\cite{ma2023diffusionseg}, but they do not consider keypoint detection. We also evaluate multi-stage pipelines for diffusion model texturing and compare with random textures on both objects and backgrounds, instead of only on the background.

\section{Synthetic Data Generation}
\label{sec:data-gen}
In this work, we generate synthetic images of static scenes containing an object on a table surface to enable generic robotic manipulation of the object category.

The data generation process consists of two steps: In the first step we gather 3D meshes, annotate these meshes and use them to generate 3D scenes. 
In the second step, we texture the scene to provide the desired visual diversity. Combining the annotations from the first step with the textured images obtained after the second step, we obtain a diverse dataset for the object category with pixel-accurate labels. 

For the second step, we compare different approaches to texture the scene, either using random textures, a common technique in domain randomization, or by using text-to-image diffusion models.
Both stages are described in more detail in the next sections. Figure~\ref{fig:overview} illustrates the data generation process used in this work.

\subsection{Scene Generation}
For each object category that we want to create synthetic data for, we first need to acquire a set of meshes. The meshes do not need to be of a very high quality and in particular do not require accurate UV-maps, which are often hard to get. 
In addition to gathering the meshes, we also need to obtain the required annotations. In this work, these are the 3D positions of the semantic keypoints and the object masks. The object masks require no additional labeling as they can be obtained from the render engine. The keypoints can be manually annotated, but it is often possible to determine them automatically based on the mesh geometry.

Once we have the meshes and the required annotations, we generate 3D scenes of the objects. To model the table surface, we simply use a 2D plane. To introduce the desired scene geometry variations, we randomize the table's dimensions as well as the object and camera pose.

In these scenes, as we know the camera intrinsics and extrinsics, we can project all annotations to the 2D images, obtaining pixel-perfect annotations. To generate visual diversity, we also need to texture the scene, which is discussed next.

\subsection{Texturing}
\label{sec:method-texturing}
To texture (an image of) the scene, we compare two different approaches: in the first approach, we simply apply random textures to the elements of the scene. In the second, we use a text-to-image diffusion model and condition it on a depth image of the 3D scene and a suitable prompt. Each approach is now discussed in more detail. 

\subsubsection{random textures}
With this method, we apply a random texture to the meshes of the object and the surface. In addition we use a 360 image as scene background to further increase visual diversity. We follow~\cite{lips2024cloth-keypoints} and use textures and 360 images from PolyHaven\footnote{\url{https://polyhaven.com/}}.

\subsubsection{diffusion texturing}
We use a depth-conditioned text-to-image diffusion model to texture the scene. More accurately, texture the image of the 3D scene.

We first generate a list of descriptions for both the object's appearance and plausible scene backgrounds. We then sample descriptions and scenes, and pass the description to the diffusion model, together with a depth image of the scene. The diffusion model then outputs an RGB image of the scene. 

By also conditioning on a depth image, we make sure that this texturing does not alter the semantics of the object, ensuring the accuracy of the predetermined scene annotations. We  use Controlnet~\cite{zhang2023controlnet} for this image conditioning and use the Stable Diffusion 1.5~\cite{rombach2022stablediffusion} text-to-image model throughout this work.

\section{Experiments}
We evaluated the data generation procedures described in previous section on three object categories: mugs, shoes and T-shirts.  We generated a dataset for each category and trained models for two object representations: keypoint detection and segmentation masks. We then reported the performance of these models on our real-world test datasets.
Section~\ref{sec:experiments-categories} provides more details about the synthetic and real datasets. The tasks and the metrics used to evaluate them are introduced in Section~\ref{sec:experiments-tasks}. The remaining sections describe the experiments we conducted, comparing diffusion-based texturing against using random textures in Section~\ref{sec:experiments-comparison} and further exploring different aspects of the diffusion-based texturing pipeline in Section~\ref{sec:experiments-additional}.

\subsection{Object categories \& datasets}
\label{sec:experiments-categories}

We evaluated three object categories: mugs, shoes, and T-shirts on a tabletop setting. For each category, we generated synthetic data using the methods described in Section~\ref{sec:data-gen}. For the mugs we gathered 100 meshes from the Objaverse~\cite{objaverse} dataset. 214 shoe meshes were obtained from the  Google Scanned Objects dataset~\cite{GSO2022google}. For the T-shirts, we used 250 meshes from~\cite{lips2024cloth-keypoints}.
For each category, 2500 distinct scenes were generated by varying the mesh pose, the size and orientation of the table and the camera pose. Fig~\ref{fig:overview} shows a number of meshes and generated 3D scenes. 5000 images were generated from these scenes by sampling different camera poses for each scene, which were then textured using one of the methods described in Section~\ref{sec:method-texturing}. We used Blender~\cite{Blender} to generate the scenes and random texture datasets. To create the diffusion textures, we used Huggingface Diffusers~\cite{diffusers}. Using an NVIDIA RTX3090 GPU, it took about 3s to render a 512x512 image with random textures using Cycles, Blender's physically-based renderer. Running inference on the diffusion model for texturing also takes around 3s per image.

We evaluated the performance on a real-world test dataset and also provided a baseline dataset with real images to put the results in perspective. For the T-shirts we used the aRTF dataset from~\cite{lips2024cloth-keypoints}, for the mugs and shoes we collected and annotated datasets manually: For the evaluation dataset we gathered a set of mugs and shoes and took pictures with a smartphone in various backgrounds. We gathered another set of mugs and shoes for the training dataset, but this time used a robot to auto-collect images from various angles. Backgrounds and objects are distinct in the train and test splits, to properly measure generalization. All images were manually annotated. The dataset sizes and number of distinct objects are given in Table~\ref{tab:real-world-statistics}. The number of objects is similar to~\cite{manuelli2019kpam}. The number of training images is about an order of magnitude smaller and more training images would likely increase the performance of the real-world baseline. Fig.~\ref{fig:overview} shows images from the real datasets on the right.

\begin{table}
    \addtolength{\tabcolsep}{-2pt}

    \centering
    \caption{Number of images and unique objects used in the real-world evaluation and baseline datasets. }
    \label{tab:real-world-statistics}
    \begin{tabular}{lcccc}
    \toprule
     &  \multicolumn{2}{c}{\textbf{train dataset} }& \multicolumn{2}{c}{\textbf{evaluation dataset}} \\
    \midrule
    \textbf{category} &  \# images & \# objects & \# images & \# objects \\
    \midrule
    Mugs & 1500& 21 &  350 & 15 \\
    Shoes & 2000& 15 & 300 & 15 \\ 
    Tshirts &  210& 15 & 400 & 20 \\

\bottomrule
    \end{tabular}

\end{table}

\subsection{Performance Evaluation}
\label{sec:experiments-tasks}
We used two different tasks to evaluate the performance of the synthetic data: semantic keypoint detection and instance segmentation. Both require precise annotations and are often used in robotics~\cite{manuelli2019kpam,vecerik2021s3k,lips2024cloth-keypoints}. For each task, we briefly discuss the training setup and the metric used to evaluate performance below. We refer to the accompanying code repository for more details.
\subsubsection{Keypoint Detection}
Following~\cite{zhou2019objects-as-points,lips2024cloth-keypoints}, we formulated keypoint detection as pixel-wise regression of target heatmaps. Each semantic category is mapped onto a different heatmap. Ground truth heatmaps are generated from the annotations by creating a Gaussian blob around each ground truth keypoint. The predicted heatmaps are regressed to the ground truth heatmaps using a binary cross entropy loss. The model is a Unet~\cite{ronneberger2015unet}-like encoder-decoder, where the encoder is a pretrained MaxVIT~\cite{tu2022maxvit} model.

To measure performance, we used the average keypoint distances (AKD) between the ground truth keypoints and
the predicted keypoint with the highest probability. 

For the T-shirts, we used the same 12 keypoints as in~\cite{lips2024cloth-keypoints}. For the shoes, we defined 3 keypoints on the nose, heel and tip.  Finally, we defined three keypoints for the mugs on the handle, bottom and top rim. These keypoints differ slightly from~\cite{manuelli2019kpam}, as we found it easier to annotate keypoints that are on the surface of the object. The keypoints are visualized in Figure~\ref{fig:overview}.

\subsubsection{Instance Segmentation}
For instance segmentation, we used YOLOv8~\cite{yolov8_ultralytics}. All hyperparameters are set at their default value and we use the small model variant, pretrained on the COCO dataset~\cite{coco}. To measure the performance, we report the mean average precision (mAP) over different IoU thresholds ranging from 0.5 to 0.95, which is the default segmentation metric for COCO~\cite{coco}.

In addition to measuring task performance of models trained on the synthetic data, which is expensive, we have tried common image metrics such as CLIP-score~\cite{hessel2021clipscore} to quantify the dataset quality, but we found these to correlate very poorly with the downstream task performance and therefore do not report them in this paper.
 
\subsection{Comparing Texturing Methods}
\label{sec:experiments-comparison}
We now compare the performance of synthetic data generated with random textures against data textured using a diffusion model, as described in Section~\ref{sec:method-texturing}.
The performance of the synthetic data generated by the different texturing methods is given in Table~\ref{tab:main-results}. We also provide the performance of a real-world train dataset as baseline. We observed that diffusion textures often perform best, though their performance was very comparable to the random textures. Finally, we observed that both random textures and diffusion textures outperform the real data baseline in most cases, confirming the efficacy of our synthetic data pipelines.


\begin{table}
    \addtolength{\tabcolsep}{-2pt}

    \centering
    \caption{Performance of the different texturing methods for all object categories and tasks. Random textures perform similar to diffusion textures. Both outperform the real baseline. }
    \label{tab:main-results}
    \begin{tabular}{lccc|ccc}
        \toprule
         & \multicolumn{3}{c}{\textbf{keypoint AKD}($\downarrow$)} & \multicolumn{3}{c}{\textbf{ seg mAP}($\uparrow$)} \\
        \cmidrule(lr){2-4}
        \cmidrule(lr){5-7}
       \textbf{Training dataset} & Mugs & Shoes & Tshirts & Mugs & Shoes & Tshirts  \\ 
        \midrule
        real data baseline & 21.7 & 33.7 & 25.6   & 0.97 & 0.88& 0.87\\ 
        \midrule
        random textures & 18.3 & \textbf{13.4} & \textbf{37.9} & 0.97 & 0.94& 0.75\\ 
        diffusion texturing &  \textbf{17.4} & 19.6 & 45.8 & \textbf{0.99} & \textbf{0.95} & \textbf{0.93} \\

\bottomrule
    \end{tabular}

\end{table}

\subsection{Further Exploration of Diffusion Texturing}
\label{sec:experiments-additional}
Next to comparing the diffusion texturing methods against random textures and a real baseline, we have performed a number of additional experiments. With these experiments we aim to validate some design choices and to provide additional insight.

We compared different strategies to generate prompts for the diffusion models, evaluated the scaling behavior of both methods, explored multi-staged diffusion pipelines, compared the Controlnet with a regular diffusion model and ablated the conditioning scale parameter used for the Controlnet. All these experiments and their results are reported in the remainder of this section.

\subsubsection{Prompting Strategy}
In this experiment we compared three different prompting strategies.

The first, and simplest strategy, is to use a fixed caption for each category, e.g., \textit{A photo of a shoe}.

To create diverse prompts and thereby more diverse images, we also used a BLIP~\cite{li2022blip} model to caption images from the real training sets for each category. We then used these captions as prompts for the diffusion model. This method aims to match the prompts to the real (target) images. We collected approximately 3000 prompts for each category using this strategy.

For the final strategy we queried an LLM (we used Google Gemini) to generate descriptions using following prompt for each object as well as for the table surfaces that are used in the scene: \textit{provide a description for X. Include color, patterns, materials and other visual characteristics.} We  randomly combined descriptions for the object and the table, obtaining a set of 5000 prompts for each category.

\begin{table}
    \addtolength{\tabcolsep}{-2pt}

    \centering
    \caption{Comparison of different prompting strategies for the one-stage data generation pipeline. Using prompts generated by an LLM produced the best results.}
    \label{tab:caption-comparison}
    \begin{tabular}{lccc|ccc}
        \toprule
         & \multicolumn{3}{c}{\textbf{AKD}($\downarrow$)} & \multicolumn{3}{c}{\textbf{mAP}($\uparrow$)} \\
        \cmidrule(lr){2-4}
        \cmidrule(lr){5-7}
       \textbf{strategy} & Mugs & Shoes & Tshirts & Mugs & Shoes & Tshirts  \\ 
       \midrule

        classname & 22.8 & 23.4 & 66.4 & 0.98 & 0.90 & 0.77 \\ 
           BLIP captions & \textbf{16.3} & 25.2 & 45.9 & \textbf{0.99} & 0.94 & 0.90 \\ 
        LLM prompts & 17.4 & \textbf{19.6} & \textbf{45.8} & \textbf{0.99} & \textbf{0.95} & \textbf{0.93}  \\ 

\bottomrule
    \end{tabular}

\end{table}

For each prompting strategy, we generated 5000 images using the one-stage diffusion texturing pipeline and trained models on these datasets for both tasks. The results are provided in Table~\ref{tab:caption-comparison}. Using a fixed template performed worse than using BLIP captions or LLM-generated prompts. The LLM-prompts scored slightly better than the BLIP captions. In addition, using LLM-prompts does not require real target images making this strategy more flexible. Therefore, we used the LLM-generated prompts in all other experiments of the paper.

Our findings are in line with~\cite{fan2023synthetic-diffusion-scaling-laws}, where the authors also found fixed templates inferior to BLIP captions. LLM-based prompts are a.o. used in~\cite{ma2023diffusionseg,ma2024generatingimages3dannotations}, but to the best of our knowledge, have not been compared explicitly against other prompting strategies for synthetic data generation.

\subsubsection{Data Scaling Behavior}
\begin{figure}
    \centering
    \includegraphics[width=0.99\linewidth]{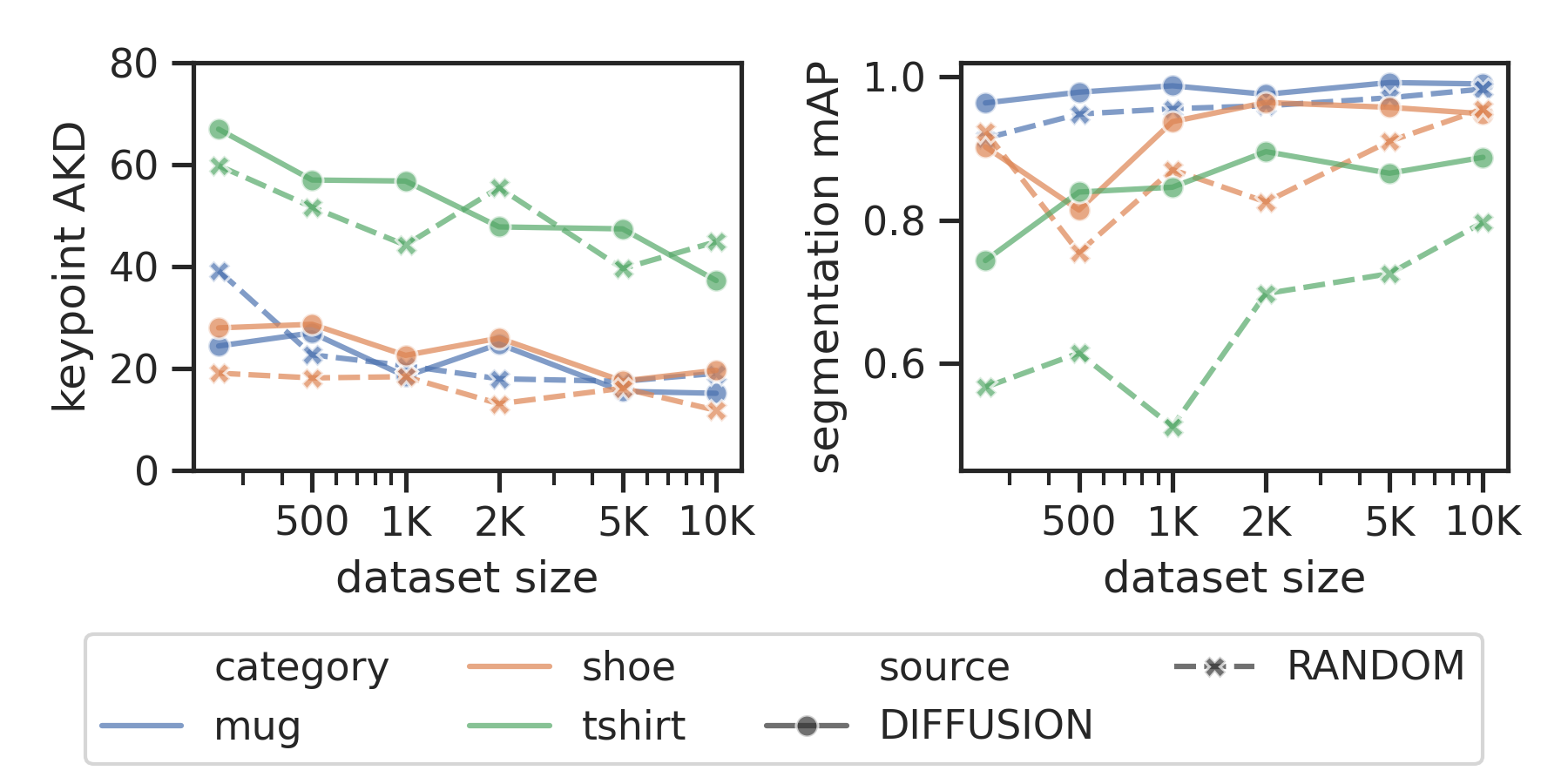}
    \caption{ Scaling behavior of the different texturing approaches. For both diffusion textures and random textures, the performance improves with increasing data, though it starts to plateau around 5,000 images.}
    \label{fig:scale}
\end{figure}

We have also explored the scaling behavior of both texturing methods. To this end, we generated a dataset with 10,000 images using both random texturing and the one-stage diffusion texturing. We then created dataset slits with various sizes and trained models for both keypoint detection and segmentation on all datasets. The performance of these models can be seen in Figure~\ref{fig:scale}. We observed that for both diffusion and random textures, the performance increased as the dataset size increases. However, around 5000 images, the performance starts to plateau, indicating that neither method was able to bridge the sim-to-real gap completely and obtain optimal performance. Based on this experiment, we have used a dataset size of 5000 for all experiments in this paper.

\subsubsection{multi-stage diffusion texturing}
\begin{figure}
    \centering
    \includegraphics[width=0.99\linewidth]{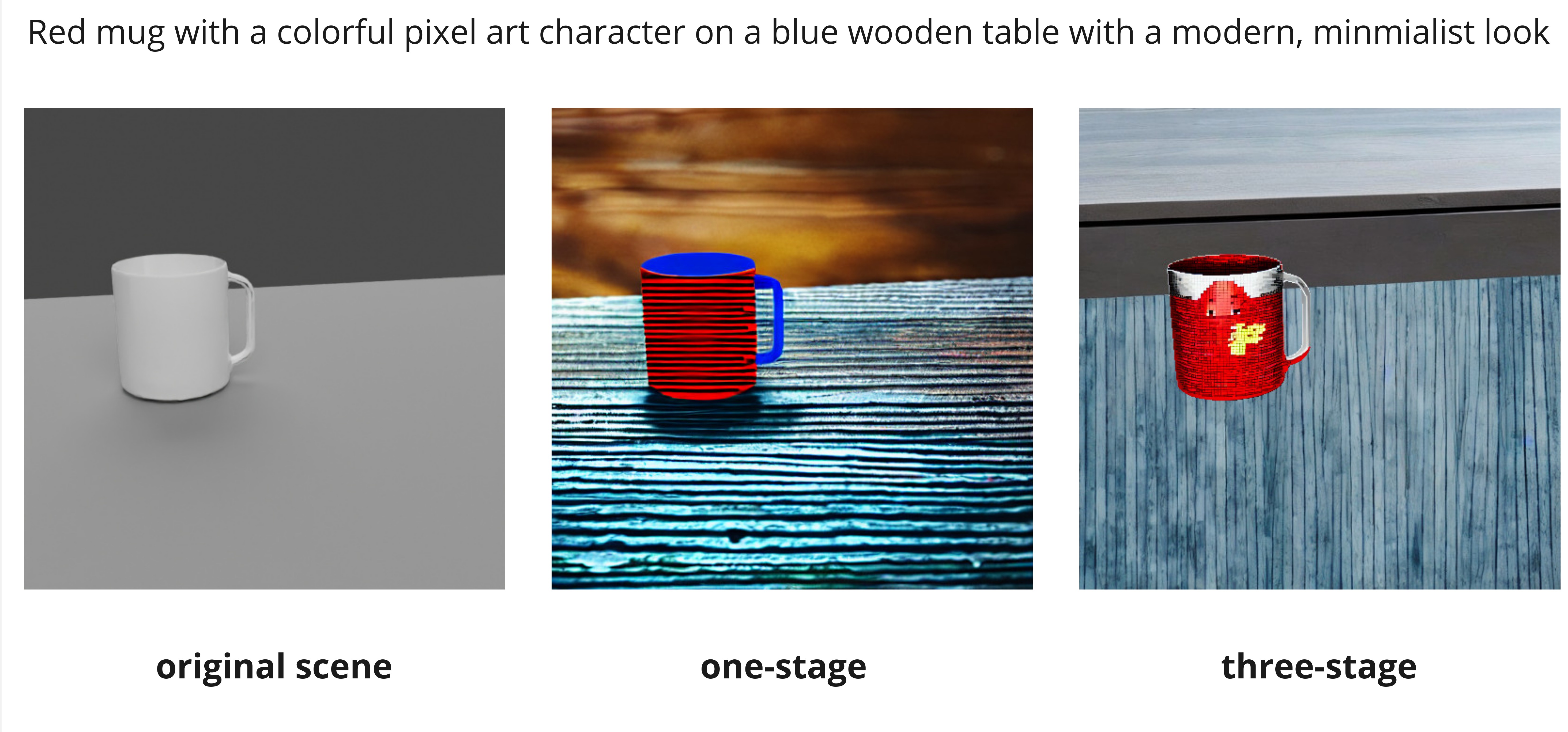}
    \caption{Illustration of the one-stage approach issue: the background and object prompt are entangled in the visual output. The three-stage approach managed to separate them (though the table perspective is wrong).}
    \label{fig:stage-comparison}
\end{figure}

We have noticed two shortcomings with the diffusion texturing approach described in section~\ref{sec:method-texturing}: the diffusion model tends to blend the description of the background with the description of the model on the one hand, and it sometimes alters the object when its apparent size in the image is small. We illustrated the former in Figure~\ref{fig:stage-comparison}. In an attempt to overcome these issues, we also explored multi-stage pipelines in which we only texture parts of the scene at each step. We describe our two- and three-step approach below, after which we compare it against the one-stage approach described in section~\ref{sec:method-texturing}.

\textbf{two-stage diffusion}
In the first stage, we condition a diffusion model on a depth image and a prompt describing the object, but we now crop the depth image to the bounding box of the object instead of the entire scene. In the second stage, we condition the diffusion model on the background text prompt to inpaint the background of the image.

Though this effectively overcomes both issues with the single-stage approach, it introduces a new issue. During the inpainting stage, the diffusion model sometimes extends the object, in an attempt to smoothen the foreground-background transition. This changes the object semantics, hence reducing data quality. To mitigate this issue, as in~\cite{chen2023genaug}, we apply a dilation to the inpainting mask to create a 'buffer'. 

\textbf{three-stage diffusion}
To completely eliminate the aforementioned issue, we also explore a three-stage approach. In this approach, we first texture the scene without the object, so with an empty table. We use the two-stage method described before, where we now crop on the table. We then crop the object mask and again condition a diffusion model on the depth image and object prompt to texture the object, which is then overlayed onto the textured image of the empty table. This should enable the model to focus on the different elements of the scene, while leaving the object semantics unchanged. Figure~\ref{fig:stage-comparison} illustrates this.

We evaluated both approaches against the one-stage approach from section~\ref{sec:method-texturing}. For each approach, we generated 5000 images using the same set of prompts and 3D scenes as in section~\ref{sec:experiments-categories}. The results are presented in table~\ref{tab:multi-stage-comparison}. We observed that the one-stage and three-stage approaches significantly outperform the two-stage approach. The three-stage approach performed similarly to the one-stage approach. However, it takes thrice as much time to produce each image. Therefore, we used the one-stage approach in the remainder of this paper.

\begin{table}
    \addtolength{\tabcolsep}{-2pt}

    \centering
    \caption{Comparison of different diffusion texturing approaches. The one-stage approach has the best overall performance. }
    \label{tab:multi-stage-comparison}
    \begin{tabular}{lccc|ccc}
        \toprule
          & \multicolumn{3}{c}{\textbf{keypoint AKD}($\downarrow$)} & \multicolumn{3}{c}{\textbf{ seg mAP}($\uparrow$)} \\
        \cmidrule(lr){2-4}
        \cmidrule(lr){5-7}
       \textbf{Approach} & Mugs & Shoes & Tshirts & Mugs & Shoes & Tshirts  \\ 
        \midrule
        one-stage  &  \textbf{17.4} & 19.6 & 45.8 & \textbf{0.99} & \textbf{0.95} & \textbf{0.93} \\
        two-stage  & 19.8 & 21.7 & 49.3 & 0.75 & 0.70 & 0.62 \\
        three-stage  & 20.9 & \textbf{19.3} & \textbf{38.3} & 0.96 & 0.91 & 0.71 \\

\bottomrule
    \end{tabular}

\end{table}

\subsubsection{Image Conditioning}
With this experiment, we aimed to validate the use of image-conditioned Controlnets~\cite{zhang2023controlnet} for data generation instead of using inpainting of object masks (which are also available from our 3D scenes) with vanilla text-to-image diffusion models as in~\cite{yu2023rosie}. We compared the one-stage diffusion method from Section~\ref{sec:method-texturing} against sequentially inpainting the object mask and background mask. The results are presented in Table~\ref{tab:inpainting vs controlnet} and confirm the necessity of Controlnet-style image conditioning. We observed that with inpainting, the diffusion model tends to alter the object semantics significantly, rendering the predetermined annotations invalid. Conditioning on (depth) images of the scenes using a Controlnet, reduces this problem.

\begin{table}
    \addtolength{\tabcolsep}{-2pt}

    \centering
    \caption{Comparison between using a Controlnet as in the one-stage pipeline from Section~\ref{sec:method-texturing} and inpainting with a Stable Diffusion model. The Controlnet significantly improves the performance.}
    \label{tab:inpainting vs controlnet}
    \begin{tabular}{lccc|ccc}
        \toprule
         & \multicolumn{3}{c}{\textbf{AKD}($\downarrow$)} & \multicolumn{3}{c}{\textbf{mAP}($\uparrow$)} \\
        \cmidrule(lr){2-4}
        \cmidrule(lr){5-7}
       \textbf{Data Pipeline} & Mugs & Shoes & Tshirts & Mugs & Shoes & Tshirts  \\ 
       \midrule
        one-stage  & \textbf{17.4 }& \textbf{19.6} & \textbf{45.8} & \textbf{0.99} & \textbf{0.95} & \textbf{0.93} \\
        inpainting & 26.3 & 48.0 & 95.2 & 0.30 & 0.42 & 0.43 \\

\bottomrule
    \end{tabular}
    \end{table}

\subsubsection{Controlnet Conditioning Scale}

\begin{figure}
    \centering
    \includegraphics[width=0.99\linewidth]{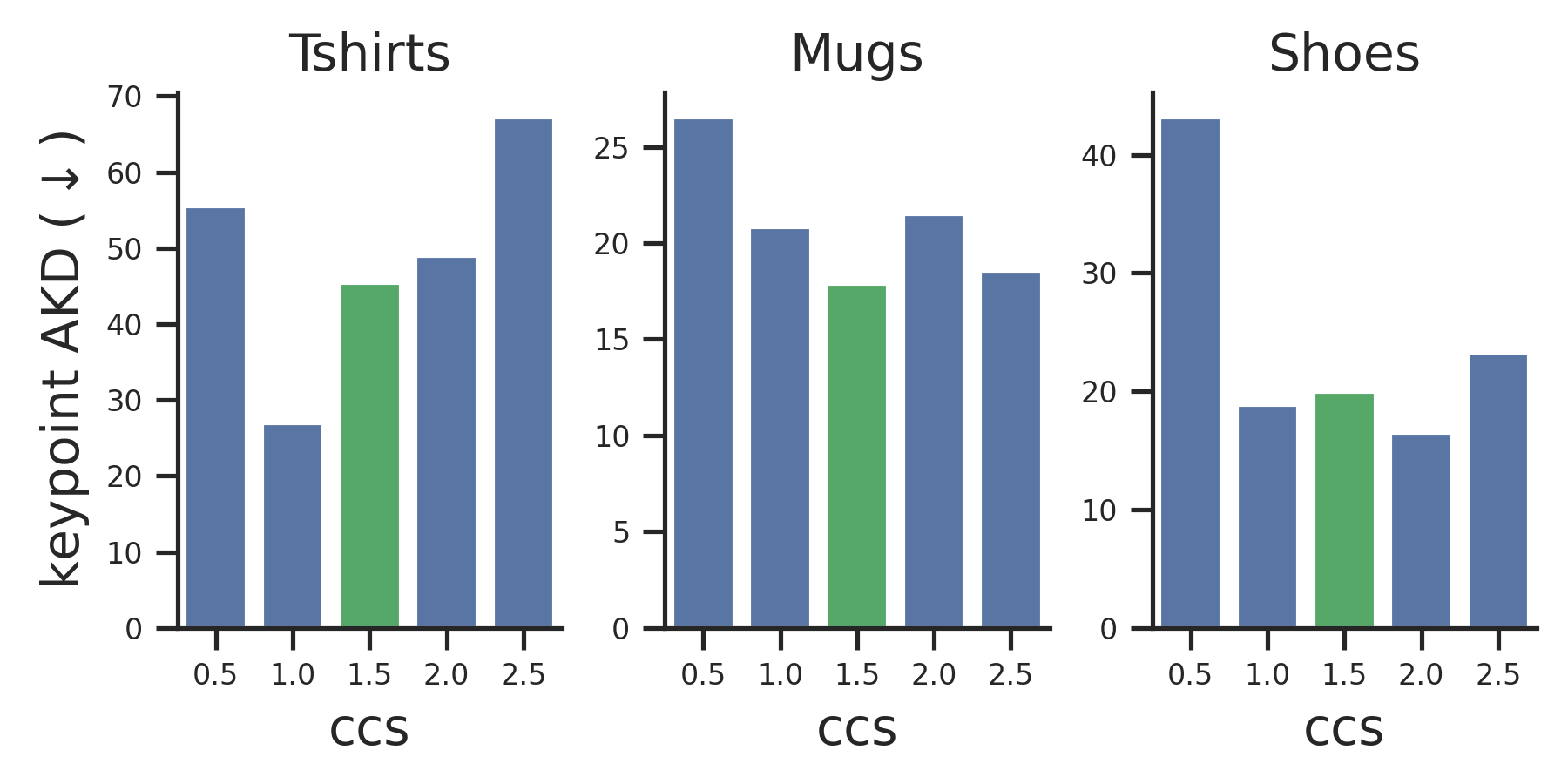}
    \caption{Comparison of average keypoint distances for different values of the Controlnet conditioning scale (CCS). The optimal value depends on the category, but 1.5 (marked in green) is a sensible default.}
    \label{fig:ccs-comparison}
\end{figure}

Controlnet~\cite{zhang2023controlnet} diffusion models,  have various hyperparameters that influence the inference process. In this paper, we set out to assess the out-of-the-box performance of diffusion models for generating synthetic data. Therefore, we have not performed rigorous hyperparameter optimizations; instead, we used the default values from the Diffusers library after validating them qualitatively. However, we found that the Controlnet conditioning scale (CCS), which determines how much importance is given to the input image (the depth image of the scene in our case), has a profound impact on the performance. Therefore we compared the performance for different values of this parameter for all categories on the keypoint detection task. The results of this experiment are shown in Fig.~\ref{fig:ccs-comparison}. The results confirmed the impact of this parameter and showed how the optimal value varies across the different categories. In this paper we wanted to use the same hyperparameters for all categories and, based on the results, chose 1.5 as default value for the remainder of the paper. To obtain optimal performance for a single object category -- which was not our goal -- this parameter should be tuned carefully though.

%
\section{Discussion}


In this work we have compared text-to-image diffusion models against random textures for texturing synthetic data. We have observed that the diffusion-based texturing pipeline does not outperform the random textures. This was surprising, as the diffusion-textured images appeared more realistic to us, and therefore we expected them to reduce the sim-to-real gap. We suspect this increased realistic appearance is countered both by the tendency of the diffusion network to slightly alter the object semantics (e.g., change the shape of the mug handle slightly), polluting the annotations, and by the diffusion models leaving strong artifacts in the synthetic images on which the models can then overfit (e.g., blurring the background). Further research is required to test these hypotheses, but there seems to be a big difference between \textit{appearing realistic} and actually matching the distribution of real-world images. 

In addition to the performance, the data generation speed is also important. We have not optimized this in our paper, the single-stage diffusion pipeline and random textures pipeline both took about 3 seconds to texture an image. Both can be sped up significantly and although diffusion models are becoming faster, we believe that the random textures pipeline will nonetheless be faster when fully optimized. 

Finally, we note that the performance of the diffusion-based pipeline strongly depends on the synthetic data context. There are limits to the semantic knowledge of a diffusion mode, imposed by the dataset it was trained on. There exist techniques to insert knowledge about new semantic categories~\cite{gal2022textual-inversion}, but these come with additional engineering and data collection effort. For diffusion-based texturing, the performance can also depend on the camera angle. We observed for example how images in which the mug handle was prominently visible tended to be more realistic than images in which the mug handle was occluded. This is in line with~\cite{ma2024generatingimages3dannotations}.

All in all, our diffusion-based texturing pipeline does not provide much gains in performance over the random texturing approach and increases complexity. At the same time, neither method scales until it has a perfect performance, so better approaches are still needed. Improving generative models, both text-to-image and text-to-3D models, is the best path to reduce engineering effort in synthetic data generation and will result in diffusion-based texturing improving random textures. End-to-end synthetic data generation, as in ~\cite{ma2023diffusionseg}, reduces complexity but requires methods to annotate the keypoints afterwards, which is even harder than for segmentation masks due to the increased precision and semantic granularity. In addition, our explicit procedure offers controllability of the generation process, allowing to further steer the data distribution.

\section{Conclusion}
In this work we evaluated the use of text-to-image diffusion models to generate synthetic data for keypoint detection and segmentation in the context of robotic manipulation. We have validated several design choices of our diffusion-based texturing pipeline in Section~\ref{sec:experiments-additional} to ensure they are appropriate and to provide insight. Surprisingly, our diffusion-based pipeline does not outperform texturing the 3D scenes using random textures, which is a conceptually simpler approach that works similarly out of the box for all objects and camera angles, unlike the diffusion pipeline. Although using generative models remains an interesting option to reduce engineering effort in synthetic data generation, it does not provide many gains for synthetic image texturing at this time.

\section*{Acknowledgments}
This project is supported by the Research Foundation Flanders (FWO) under Grant numbers 1S56022N and by the euROBIn Project (EU grant number 101070596). The authors also thank Victor-Louis De Gusseme for his input.
\bibliographystyle{IEEEtran}
\bibliography{references}

\end{document}